\documentclass[sigconf,authordraft]{acmart}
\renewcommand\footnotetextcopyrightpermission[1]{} 
\makeatletter
\def\runningfoot{\def\@runningfoot{}}
\def\firstfoot{\def\@firstfoot{}}
\makeatother 

\usepackage{amssymb}
\usepackage{amsmath}
\usepackage{cleveref}
\usepackage{booktabs}
\newcommand{\ra}[1]{\renewcommand{\arraystretch}{#1}}
\AtBeginDocument{%
  \providecommand\BibTeX{{%
    \normalfont B\kern-0.5em{\scshape i\kern-0.25em b}\kern-0.8em\TeX}}}

\setcopyright{none}

\begin{document}

\title[ ]{Adaptive Margin Ranking Loss for Knowledge Graph Embeddings via a Correntropy Objective Function}

\author{Mojtaba Nayyeri}
\affiliation{%
  \institution{University of Bonn}
  \city{Bonn}
  \country{Germany}}
\email{nayyeri@cs.uni-bonn.de}

\author{Xiaotian Zhou}
\affiliation{%
  \institution{University of Bonn}
  \city{Bonn}
  \country{Germany}}
\email{Zhou@cs.uni-bonn.de}

\author{Sahar Vahdati}
\affiliation{%
  \institution{University of Bonn}
  \city{Bonn}
  \country{Germany}}
\email{vahdati@cs.uni-bonn.de}

\author{Hamed Shariat Yazdi}
\affiliation{%
  \institution{University of Bonn}
  \city{Bonn}
  \country{Germany}}
\email{shariat@cs.uni-bonn.de}

\author{Jens Lehmann}
 \affiliation{%
  \institution{University of Bonn, Fraunhofer IAIS}
  \city{Bonn}
  \country{Germany}}
\email{jens.lehmann@cs.uni-bonn.de}

%

\renewcommand{\shortauthors}{Nayyeri et al.}

%


%

%

\begin{abstract}
Translation-based embedding models have gained significant attention in link prediction tasks for knowledge graphs. 
TransE is the primary model among translation-based embeddings and is well-known for its low complexity and high efficiency. 
Therefore, most of the earlier works have modified the score function of the TransE approach in order to improve the performance of link prediction tasks. 
Nevertheless, proven theoretically and experimentally, the performance of TransE strongly depends on the loss function.
Margin Ranking Loss (MRL) has been one of the earlier loss functions which is widely used for training TransE. 
However, the scores of positive triples are not necessarily enforced to be sufficiently small to fulfill the translation from head to tail by using relation vector (original assumption of TransE). 
To tackle this problem, several loss functions have been proposed recently by adding upper bounds and lower bounds to the scores of positive and negative samples. 
Although highly effective, previously developed models suffer from an expansion in search space for selection of the hyperparameters (in particular the upper and lower bounds of scores) on which the performance of the translation-based models is highly dependent.
In this paper, we propose a new loss function dubbed Adaptive Margin Loss (AML) for training translation-based embedding models.
The formulation of the proposed loss function enables an adaptive and automated adjustment of the margin during the learning process.
Therefore, instead of obtaining two values (upper bound and lower bound), only the center of a margin needs to be determined. 
During learning, the margin is expanded automatically until it converges. 
In our experiments on a set of standard benchmark datasets including Freebase and WordNet, the effectiveness of AML is confirmed for training TransE on link prediction tasks.

\end{abstract}

\maketitle 

\section{Introduction}

Knowledge graphs are one of the
most important technologies for the next wave of artificial intelligence and knowledge management solutions across industrial applications~\cite{bini2018artificial,adams2019surfing,panetta5trends}.
This is evident by a broad range of use cases of KGs ranging from question answering~\cite{bordes2014question,he2014question,swj_survey_qa}, recommendation systems~\cite{zhang2016collaborative}, semantic modeling~\cite{shen2013linking} to data analysis~\cite{lin2016neural}, and knowledge management systems~\cite{chen2016neural,szumlanski2010automatically}.
To support such intelligent applications, various large-scale knowledge graphs have been made available. 
Some of the most used knowledge graphs are WordNet~\cite{miller1995wordnet}, Freebase~\cite{bollacker2008freebase}, NELL~\cite{carlson2010toward}, Yago~\cite{nickel2012factorizing} and DBpedia~\cite{swj_dbpedia}.
These datasets include knowledge in multi-relational directed graphs composed of nodes $\mathcal{E}$ (usually called entities) and edges $\mathcal{R}$ (usually called links or relations).
More precisely, a $\mathcal{KG}$ includes a set of triples in the form of (head, relation, tail) denoted as $(h,r,t)$ where $h, t$ refer to the subject (also called head) and object (also called tail) respectively and $r$ refers to a relation. 
This representation of information empowers navigation across information and provides an effective utilization of encoded knowledge. 
Since it is difficult to capture all the existing knowledge from the real world, knowledge graphs are usually incomplete. 
This limits the inference of knowledge and influences performance of the systems utilizing such KGs.  
An elegant solution to solve the incompleteness of KGs are ``Knowledge Graph Embeddings (KGE)''. 
Those embeddings assign a latent feature vector to each node and relation in a KG, which can then be used in downstream machine learning tasks such as link prediction. 
\begin{figure*}[ht!]
\centering
\includegraphics[trim=.3cm 2cm 0cm 1cm,clip,width=1.0\textwidth]{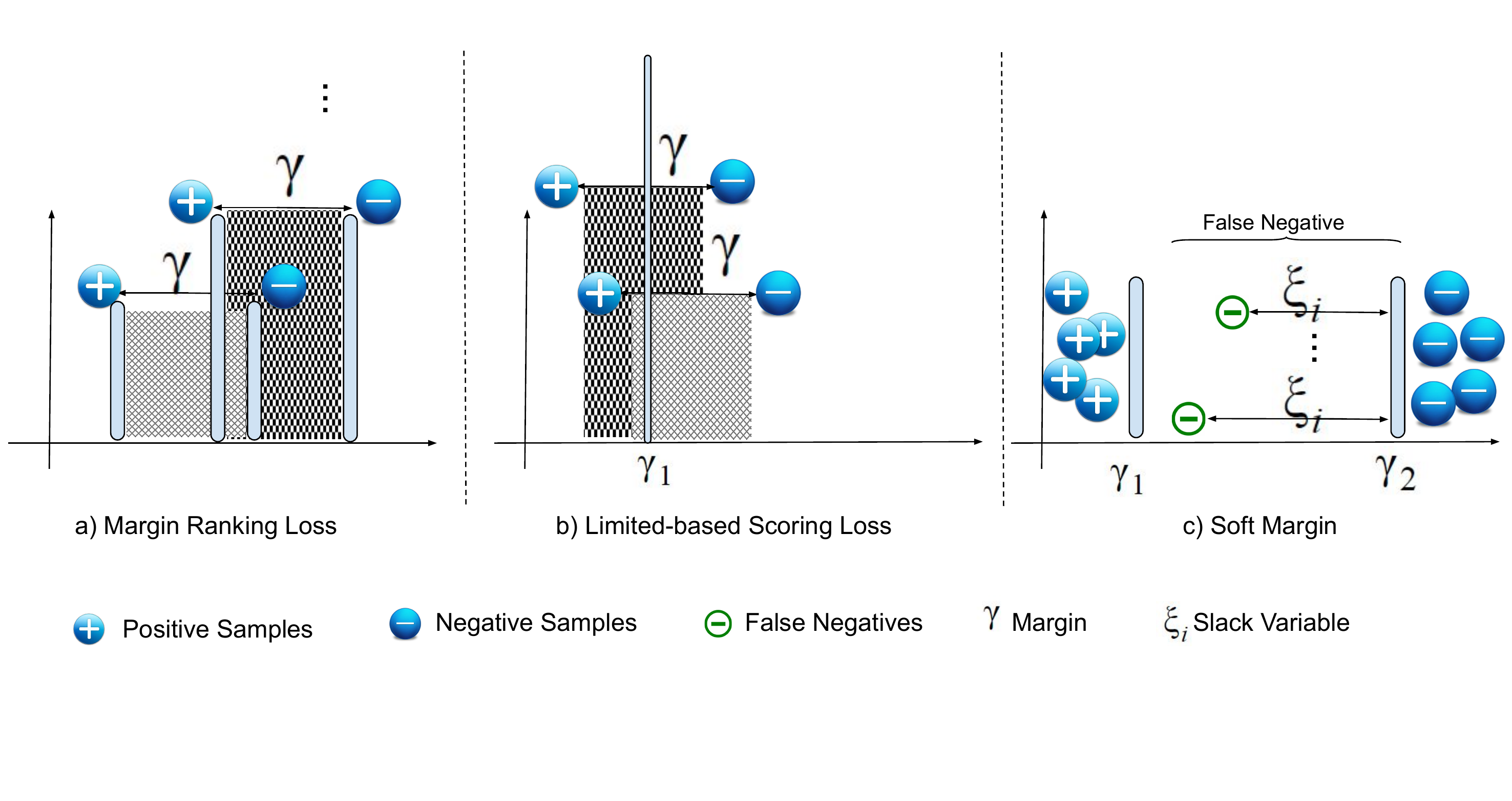} 
\caption{Illustration of Loss Functions.}
	\label{fig:sota}
\end{figure*}
Among the proposed KGE methods, translation-based models are considered as a key family of methods for graph completion tasks.  
Translation-based models encode entities as vectors and relationships between entities as translation vectors.
TransE~\cite{bordes2013translating} is one of the primary models that seeks for a latent feature vector representation of a given triple $(h,r,t)$ in which the vector representing $t$ is same as the sum of the vectors representing $h$ and $r$.
Initially, the corresponding vectors ($\textbf{h}, \textbf{r}$, $\textbf{t}$) of each individual triple ($h$, $r$, $t$) are randomly distributed over the vector space.
An embedding model employs a scoring function and a loss function in order to (approximately) satisfy $\textbf{h} + \textbf{r}  \simeq \textbf{t}$ for positive triples $(h,r,t)$, and  $\textbf{h} + \textbf{r} \neq \textbf{t}$ for negative samples of $(h',r,t')$.
The correctness of a $(h,r,t)$ triple is calculated via a scoring function in the embedding space such as  $  f_r(h,t) = \| \textbf{h} + \textbf{r} - \textbf{t} \| $.
Since the vectors for positive and negative (corrupted) triples are randomly distributed, the results of the scoring function also evaluates their correctness randomly as well.
Therefore, a loss function (e.g.\ the Margin Ranking Loss) is needed to optimize the embedding vectors of entities and relations.

MRLs are widely accepted and used in embedding models and their effectiveness is proven~\cite{trouillon2016complex, bordes2013translating}.
The margin-based ranking loss function forces the score of positive triples to be lower (towards $0$) and assigns a higher score to negative triples by a margin of at least $\gamma$. Therefore, positive triples are separated from negative samples. 
However, using MRL includes the existence of cases where the score of a correct triple $(h,r,t)$
is not sufficiently small for $\textbf{h} + \textbf{r}  \simeq \textbf{t}$ to hold.
A combination of limit-based scoring loss functions for a set of translation-based embedding models~\cite{zhou2017learning} have been proposed in order to avoid such cases.
By adding a limit of $ f_r(h, r) \leq \gamma_1$, the score of correct triples is bounded within a determined range. 
However, the setting of $\gamma_1$ and $\gamma_2$ in alignment with the score of the positive and negative triples is done with a ``trial and error'' method in a very big search space. 
Due to the lack of a unique answer for sliding $\gamma_1$ (upper-bound of positive triples) and $\gamma_2$ (lower-bound of negative triples) and the large search space, this task can be multiplied for any possible variation in ranking. 
In this work, we propose an adaptive margin loss function for translation-based embedding models. 
Our method reduces the search of two hyperparameters ($\gamma_1, \gamma_2$) to one variable ($\gamma$).
$\gamma$ is the center of the margin that should be searched within a set of numbers. 
The margin is adjusted automatically during the learning process by formulating a slack variable in the optimization problem. 

The remaining part of this paper proceeds as follow.
Section \ref{relatedwork} represents the related work and previous proposals developed for loss functions of Translation-based embeddings. 
Section \ref{sec:adamtiv} provides a detailed description of the adaptive model.
An evaluation of the newly developed loss function is shown in Section \ref{sec:expr}.
In Section \ref{sec:concl}, we lay out the insights and provide conjunction of this research work.

\section{Related Work}\label{relatedwork}
The loss function has a significant impact on the performance of translation-based embedding models~\cite{zhou2017learning,anonymous2018relation}.
Defining a margin to separate positive and negative triples is one of the promising solutions in keeping a high performance for loss functions. 
Therefore, approaches focusing on a proper adjustment for such a margin in the loss function became an important task in translation-based KGEs. 
Here, we introduce three of the main proposed margin-based ranking loss functions. 
An illustration of each loss function is shown in \autoref{fig:sota}.

\subsection{Margin Ranking Loss} 
Margin Ranking Loss (MRL) is one of the primary approaches that was proposed to set a margin of $\gamma$ between positive and negative samples.
It is defined as follows:
\begin{align}
    \mathcal{L} = \sum_{(h,r,t) \in S^{+} }^{~} \sum_{(h',r',t') \in S^{-} }^{~}\,
    [f_r(h,t) + \gamma - f_r(h', t')]_+
\end{align}
where $[x]_+=\max(0,x)$ ($S^{+}$ for positive samples and $S^{-}$ for negative samples).
$S^{-}$ includes training samples with two patterns of triples: 1) a corrupted head replaced by a random entity for a fixed tail , 2) for a fixed head, a corrupted tail is replaced by a random entity. 
The score of any such corrupted triple $f_r(h',t')$ in negative samples is forced to be higher than the positive triples $f_r(h,t)$ with a margin of $\gamma$. 
The loss function assigns scores to the parameters in a way that $ f_r(h',t') - f_r(h, r) \geq \gamma$ holds. 
However, this loss function does not guarantee that the scores assigned to the positive samples are low enough to present the correct translation (i.e.\ $\textbf{h} + \textbf{r} \simeq \textbf{t}$).
It is possible that the model forcing to hold this condition assigns scores for the following positive and negative samples (for an initial $\gamma = 1$):
\begin{equation}
     \begin{cases}
   (f_r(h',t') = 1) - (f_r(h, r) = 0) \geq (\gamma = 1)
   \\
    (f_r(h',t') = 11) - (f_r(h, r) = 10) \geq (\gamma = 1)
   \\
    (f_r(h',t') = 101) - (f_r(h, r) = 100) \geq (\gamma = 1)
    \\
    (f_r(h',t') = 1001) - (f_r(h, r) = 1000) \geq (\gamma = 1)
     \end{cases}
     \label{examples}
 \end{equation}
Although the calculated loss is the same number for each of these examples, the score scale of the latter sample is higher than the first one. 
This makes the positive training triples with high scores hardly meeting the conditions of $\textbf{h} + \textbf{r}  \simeq \textbf{t}$, illustrated as Margin Ranking Loss in \autoref{fig:sota}.
Thus, with such a loss function, it is possible that the model produces ineffective results. 

\subsection{Limited-based Scoring Loss} 
In order to fulfill the gap of MRL in assigning high scores to positive samples, a limited-based scoring function has been proposed~\cite{zhou2017learning}.
This method limits the score of positive samples by adding an upper-bound ($\gamma_1$).
It is represented as limited-based scoring loss illustrated in \autoref{fig:sota}.
In this way, the scores of positive samples are forced to stay before the upper bound which significantly improves the performance of translation-based KGE models~\cite{zhou2017learning,nayyeri2019soft,anonymous2018relation}.
\citet{zhou2017learning} revises the MRL by adding a term ($[f_r(h,t) - \gamma_1]_+$) to limit maximum value of positive score: 
\begin{align}
    \mathcal{L}_{RS} = \sum \sum \,
    [f_r(h,t) + \gamma - f_r(h', t')]_+ + \lambda [f_r(h,t) - \gamma_1]_+
\end{align}
The possible combination of variables for $\gamma$ and $\gamma_1$ is wide with a complexity of $O(n^2)$.
Considering that, the setting of ($\gamma, \gamma_1$) is yet a manual task in experiments, the model and the results suffer from the difficulty of finding an optimum setting by trying all possible combinations.

\subsection{Soft Margin} 
A modified version of the two previous loss functions is introduced in our previous work\cite{nayyeri2019soft}.
This approach fixes the upper-bound of positive samples ($\gamma_1$) and uses a sliding mechanism to move false negative samples towards positive samples, shown as Soft Margin in \autoref{fig:sota}. 
$\theta$ refers to embedding parameters of all entities and relations in KG as (\textbf{h}, \textbf{r}, \textbf{t}). 
A slack variable is used per each triple (i.e.\ $\xi_i$, where $i$ refers to the $i$-th triple) to enable false negative samples to slide inside the margin. 

\begin{equation} 
\begin{split}
\min_{\xi_{h,t}^r, \theta} \sum_{(h,r,t) \in S^+} \lambda\, {\xi_{h,t}^r}^2 + \lambda_+ [f_{r}(h,t) - \gamma_1]_+ \, + \\ \lambda_-\, [\gamma_2 - f_{r}(h',t') - {\xi_{h,t}^r}]_+
\end{split}
\label{uncOpt}
\end{equation}

In order to properly adjust margin, two variables ($\gamma_1, \gamma_2$) should be obtained. 
Experiments show that the performance of KGE models improves significantly by using different values for ($\gamma_1, \gamma_2$). 
Assuming $\gamma_1$ in the range of 10 possible variables $\{0, 0.5, 1, \dots, 4.5\}$ and $\gamma_2$ in another range of 10 possibilities such as $\{0.5,1,1.5, \dots, 5\}$ result in $10^2$ variations for ($\gamma_1, \gamma_2$). 
The setting of ($\gamma_1, \gamma_2$) is yet a manual task in experiments, the model and the results suffer from the difficulty of finding an optimum setting by trying all possible combinations. 
The results are promising with a focus on handling uncertainty in negative sampling (false negative samples).
However, a correct setting of $\gamma_2$ in alignment with $\gamma_1$ still remains challenging for the performance and effectiveness of the model. 

\section{Adaptive Margin Ranking Loss}
\label{sec:adamtiv}
\begin{figure*}[ht!]
\centering
\includegraphics[trim=.3cm 2cm 0cm 1cm,clip,width=1.0\textwidth]{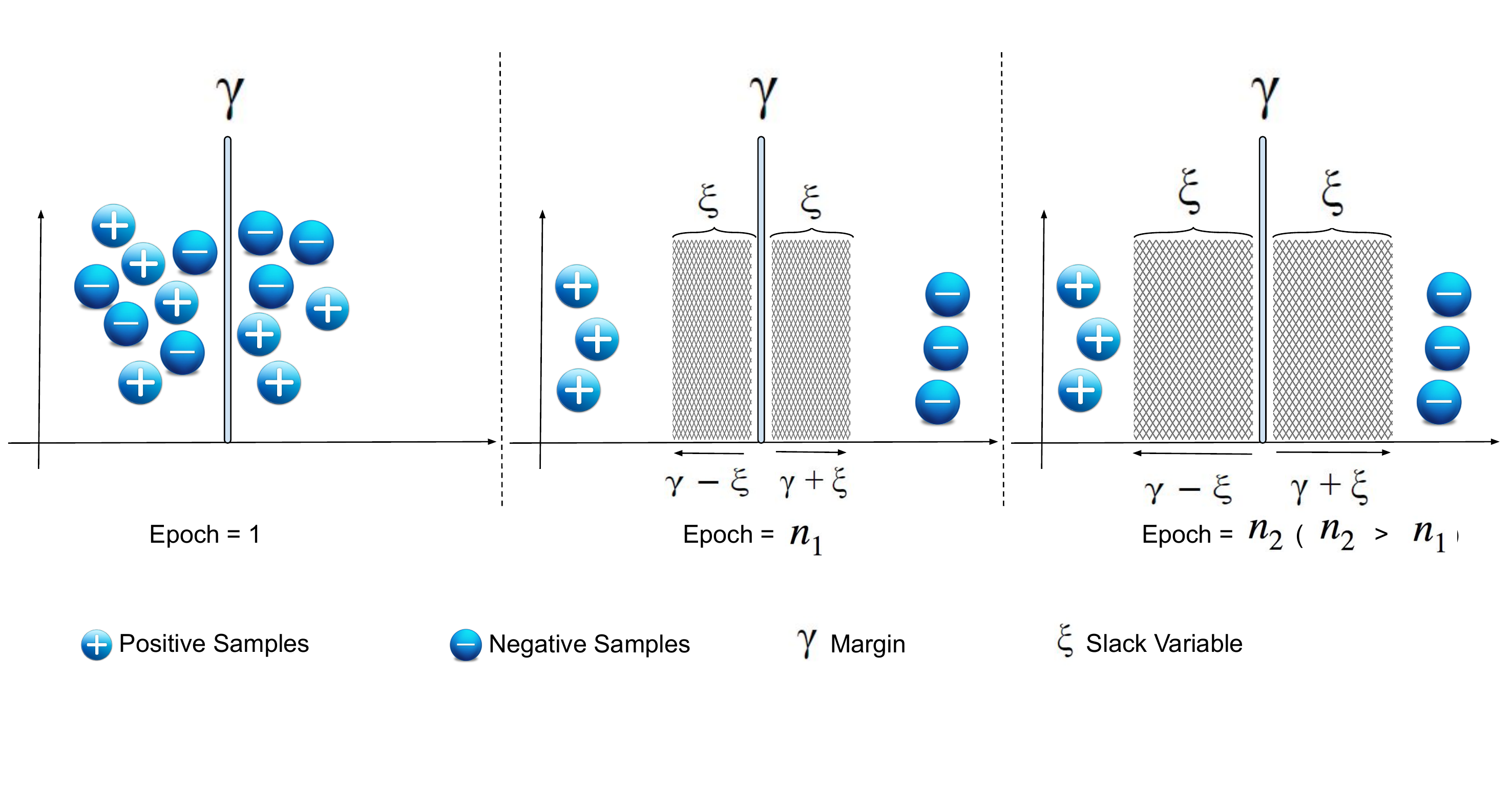} 
\caption{Illustration of Adaptive Margin Loss.}
	\label{fig:adaptive}
\end{figure*}

Inspired by MRL and specifically aiming at the reduction of search space, we use a variable ($\gamma$) denoting the center of the margin between positive and negative scores.
As a result, instead of searching for two parameters ($\gamma_1, \gamma_2$), we search for one parameter ($\gamma$) illustrated in \autoref{fig:adaptive}. 
We propose two separate loss functions to obtain the margin automatically.
One of the loss functions is using \emph{expansion} approach (denoted by $\mathcal{L}_E$) and the other uses \emph{contraction} (denoted by $\mathcal{L}_C$).
The \textbf{expansion} method gradually increases the margin from zero to a bigger value. 
In the other method, for \textbf{contraction}, the margin shrinks from bigger values to smaller ones.
These two methods are independent and are for solitary usage.
The performance of each method depends on the application area and the general status of the KG and the underlying model.
The authors leave the decision of using contraction or expansion methods on users based on the best performance of each loss in the defined embedding problem. 

A slack variable ($\xi$) is employed to gradually expand (or contract) the margin i.e.\ $\gamma_1 = \gamma - \xi, \gamma_2 = \gamma + \xi$. 
Therefore, the following inequalities should hold for positive and negative scores: 
 
\begin{equation}
 \begin{cases}
    f_r(h,t) \leq \gamma - \xi, \\
    f_r(h^{'},t^{'}) \geq \gamma + \xi. 
    \label{eq:3}
 \end{cases}
\end{equation}

Instead of using one slack variable per triples (as it was in Soft margin), we propose to use one slack variable to adapt the margin by expansion or contraction. 
In order to enforce the model to satisfy \autoref{eq:3}, the following penalty terms are derived to be included in the proposed optimization problem. 
Therefore, the loss functions of positive and negative samples are derived as follows:

\begin{equation}
 \begin{cases}
    Loss^{+} = [f_r(h,t) - \gamma + \xi]_+ = Relu(f_r(h,t) - \gamma + \xi) \\
    Loss^{-} = [- f_r(h^{'},t^{'}) + \gamma + \xi]_+ = Relu(- f_r(h^{'},t^{'}) + \gamma + \xi). 
    \label{eq:4}
 \end{cases}
\end{equation}

The initial formulation of the optimization problem is as follows:

\begin{equation}
    L = \lambda_{+}\, Loss^{+} + \lambda_{-}\, Loss^{-} 
    \label{eq:5}
\end{equation}

The role of $\xi$ is to derive the margin. 
It is initialized in the beginning of the algorithm, $\xi = 0$ for expansion ($\xi = \mathcal{M}$ for contraction). 
The initial value of the margin is introduced in \autoref{eg:margin}:
\begin{equation}
\gamma_2 - \gamma_1 = \gamma + \xi - \gamma + \xi = 2 \xi = 0 \,(2\mathcal{M}).
\label{eg:margin}
\end{equation}

\subsection{Contraction Approach}
In the contraction approach, the loss is formulated in such a way that the margin starts with a big value and gradually shrinks. 
In order to formulate the loss function with contracted margin, the following formula is employed to be added to $L$ (\autoref{eq:5}):

\begin{equation}
    L_{\xi} = \xi^2.
    \label{xi2}
\end{equation}

Therefore, considering \autoref{xi2} and \autoref{eq:3}, the following optimization is proposed:

\begin{equation}\label{eq:optimization}
\begin{aligned}
&\min_{\theta, \xi} \, \xi^2  \\
&\text{subject to} & f_r(h,t) \leq \gamma - \xi^2, \\   &&\,f_r(h^{'},t^{'}) \geq \gamma + \xi^2 \,.
\end{aligned}
\end{equation}
As explained previously, the variable $\theta$ denotes embedding parameters.

Adding a penalty parameter multiplied by a measure of violation of constrains is a solution to solve such constrained problems~\cite{boyd2004convex}.
Using penalty method (by adding $[f_r(h,t) - \gamma + \xi]_{+} + \lambda_{-}$ and $ [- f_r(h^{'},t^{'}) + \gamma + \xi]_{+}$) and considering \autoref{eq:5} and \autoref{xi2}, instead of solving \autoref{eq:optimization}, the following loss function is minimized:

\begin{equation}
    \mathcal{L} = \lambda \xi^2 + \lambda_{+} \, [f_r(h,t) - \gamma + \xi]_{+} + \lambda_{-} \, [- f_r(h^{'},t^{'}) + \gamma + \xi]_{+}
    \label{LossCont}
\end{equation}

The algorithm starts with a value for $\xi$, i.e.\ $\mathcal{M}$. 
Because the loss \autoref{LossCont} is minimized, $\xi^2 \xrightarrow{} m$ where $m < \mathcal{M}$. 
Therefore, the margin shrinks from $2M$ to $2m$. 

\subsection{Expansion Approach}
In the expansion approach, the margin is initialized with a very small value (e.g.\ zero).
Then during the optimization process, margin expands automatically.
We employ \textit{correntropy objective function} to enable the margin to be expanded.
This step is done by increasing the value of $\xi$.
The correntropy objective function is defined as follows~\cite{liu2007correntropy}:
\begin{equation*}
    \mathcal{C}(\xi) = E(K(\xi)),
\end{equation*}
where $E(.)$ is the expectation in probability theory, $K(.)$ is a kernel function and $\xi \in R^d$ is a $d-$dimensional random variable. 
Typically, Gaussian kernels are used in the correntropy function. A Gausian kernel is defined as well:

\begin{equation*}
    K(\xi) = e^{-\sigma \|\xi\|^2}.
\end{equation*}
Assuming $\xi \in R,$ which is a number rather than a vector, the following part is added to the loss $L$ (\autoref{eq:5}):

\begin{equation}
L_{\xi} = e^{-\sigma \xi^2}.
\label{eq:6}
\end{equation}

In the original \autoref{eq:3} on which the expansion will be formulated, $\xi$ should be a positive value. 
In order to ensure this, 
we use $\xi^2$ instead of $\xi$ in the final formulation of the loss function:

\begin{equation}\label{mainopt}
\begin{aligned}
  &  min_{\theta, \xi} \,\, \, e^{-\sigma \xi^2} & \\
  & \text{subject to} &   f_r(h,t) \leq \gamma - \xi^2, \\   && \,f_r(h^{'},t^{'}) \geq \gamma + \xi^2 \,.
\end{aligned}
\end{equation}

Using penalty method and considering \autoref{eq:5}, \autoref{eq:6}, instead of solving \autoref{mainopt}, the following loss function is minimized:

\begin{equation}
    \mathcal{L} = \lambda e^{-\sigma\xi^2} + \lambda_{+} \, [f_r(h,t) - \gamma + \xi]_{+} + \lambda_{-} \, [- f_r(h^{'},t^{'}) + \gamma + \xi]_{+}
    \label{lossExp}
\end{equation}

\begin{table*}\centering
\ra{1.3}
\begin{tabular}{@{}rrrrcrrcrrcrr@{}}\toprule
 \multicolumn{1}{l}{Dataset} &  & \multicolumn{3}{c}{WN18} &  &  &  \multicolumn{5}{c}{FB15k} &  \\\midrule
 & \multicolumn{2}{c}{Mean} & \phantom{abc}& \multicolumn{2}{c}{Hits@10(\%)} &
\phantom{abc} & \multicolumn{2}{c}{Mean} & \phantom{abc} & \multicolumn{2}{c}{Hits@10(\%)}\\
\cmidrule{2-3} \cmidrule{5-6} \cmidrule{8-9} \cmidrule{11-12}
  & raw & filter  && raw & filter  && raw & filter && raw & filter \\\midrule
\multicolumn{1}{l}{Unstructured \cite{bordes2012joint}}     & 315         & 304        &&   35.5          & 38.2      &&  1074        &  979       &&  4.5     &   6.3            \\
\multicolumn{1}{l}{RESCAL \citet{nickel2012factorizing}}    & 1180        & 1163       &&   37.2          & 52.8      &&  828         &  683       &&  28.4    &   44.1           \\
\multicolumn{1}{l}{SE \cite{bordes2011learning}}            & 1011        & 985        &&   68.5          & 80.5      &&  273         &  162       &&  28.8    &  39.8           \\
\multicolumn{1}{l}{SME (linear) \cite{bordes2012joint}}     & 545         & 533        &&  65.1           & 74.1      &&  274         &  154       &&  30.7    &  40.8           \\
\multicolumn{1}{l}{SME (bilinear) \cite{bordes2012joint}}   & 526         & 509        &&  54.7           & 61.3      &&  284         & 158        &&  31.3    &  41.3           \\
\multicolumn{1}{l}{LMF \cite{jenatton2012latent}}           & 469         & 456        &&  71.4           & 81.6      &&  283         & 164        &&  26.0    &  33.1           \\
\multicolumn{1}{l}{TransE \cite{bordes2013translating}}     & 263         & 251        &&  75.4           & 89.2      &&  243         & 125        &&  34.9    &  47.1           \\
\multicolumn{1}{l}{TransH (unif) \cite{wang2014knowledge}}  & 318         & 303        &&  75.4           & 86.7      &&  211         & 84         &&  42.5    &  58.5           \\
\multicolumn{1}{l}{TransH (bern) \cite{wang2014knowledge}}  & 401         & 388        &&  73.0           & 82.3      &&  212         & 87         &&  45.7    &  64.4           \\
\multicolumn{1}{l}{TransR (unif) \cite{lin2015learning}}    & 232         & 219        &&  78.3           & 91.7      &&  226         & 78         &&  43.8    & 65.5           \\
\multicolumn{1}{l}{TransR (bern) \cite{lin2015learning}}    & 238         & 225        && 79.8            & 92.0      &&  198         & 77         &&  48.2    & 68.7           \\
\multicolumn{1}{l}{TransD (unif) \cite{ji2015knowledge}}    & 242         & 229        &&  79.2           & 92.5      &&  211         & 67         &&  49.4    & 74.2           \\
\multicolumn{1}{l}{TransD (bern) \cite{ji2015knowledge}}    & 224         & 212        &&  79.6           & 92.2      &&  194         & 91         &&  53.4    & 77.3           \\ 
\multicolumn{1}{l}{TransE-RS(unif) \cite{zhou2017learning}} & 362         & 348        &&  80.3           & 93.7      &&  161         & 62         && 53.1     & 72.3           \\
\multicolumn{1}{l}{TransE-RS(bern) \cite{zhou2017learning}} & 385         & 371        &&  80.4           & 93.7      &&  161         & 63         &&  53.2    & 72.1           \\
\multicolumn{1}{l}{TransH-RS(unif) \cite{zhou2017learning}} & 401         & 389        &&  81.2           & 94.7      &&  163         & 64         &&  53.4    & 72.6           \\
\multicolumn{1}{l}{TransH-RS(bern) \cite{zhou2017learning}} & 371         & 357        &&  80.3           & 94.5      &&  178         &  77        &&  53.6    & 75.0           \\ \midrule
\multicolumn{1}{l}{TransEAML}                             & 226         & 217        &&  \textbf{84.3}           & \textbf{95.2}      &&  \textbf{159}         & \textbf{49}         &&  \textbf{55.5}    & \textbf{77.8}           \\  
\bottomrule
\end{tabular}
\caption{\textbf{Link prediction results.} Comparison of models implemented with loss function of MRL, Limited-base loss, and adaptive margin loss considering Mean rank, Hits@10 on WM18 and FB15k.}
\label{tab:1}
\end{table*}

By initializing $\xi$ to $0$, the amount of loss in \autoref{eq:6} becomes $1$ (it is maximized). 
The minimization of the main loss (equation \autoref{lossExp}) is realized when $\xi$ is enforced to be increased.
In theory this happens when $e^{-\xi^2} \xrightarrow{} 0$ which holds when $-\xi^2 \xrightarrow{} -\infty$. 
In practice, we have solved the optimization using stochastic gradient descent where $\xi$ is enforced to reach a big value ($\mathcal{M}$). 
Therefore, as indicated in \ref{eg:margin}, the margin is expanded from $0$ to $2\mathcal{M}.$ 
We emphasize on substitute usage of the \emph{expansion} and \emph{contraction} methods per use case. 
The performance of each method can differ in various applications and can be selected based on best performance. 

\section{Experiments}
\label{sec:expr}
An evaluation of our proposed adaptive margin ranking loss function is addressed in this section.
We mainly focused on training the TransE model with the state-of-the-art loss functions and provided comparisons with adaptive margin ranking loss.
The main evaluation metrics for link prediction tasks are Mean Rank (MR) and Hit@K.
To compute MR, two sets are generated ($S_L = {(h,r,?)}, S_R = {(?,r,t)}$) for each test triples ($h,r,t$) where all entities in the KGs are replaced by $?$. 
Scores of all triples in $S_L, S_R$ are computed and sorted. 
The rank of the original triple (i.e.\ $(h,r,t)$) is computed in both sets $S_L, S_R$ which are respectively denoted by $r_L, r_R$. 
In any considered triple, $r_L$ is the notation for the left ranks and $r_R$  for the right ranks. 
The rank of the example triple of ($h,r,t$) is computed as $r = \frac{r_L + r_R}{2}$. 
In this way, MR is obtained by taking overall average rank of testing triples. 
Finally, the computation of Hit@10 is performed by counting the number of testing triples which are ranked less than $10$ (i.e.\ $r_i \leq 10$).

\begin{table}[h!]
\begin{tabular}{@{}|l|l|l|l|l|@{}}
\toprule
Data          & FB15k      & wn18       \\ \midrule
Optimizer     & Adagrad       & Adagrad        \\
Embedding size & 100             & 100                \\
Epochs        & 900000    & 900000         \\
$\xi$   & 0.1              & 0.1               \\
$\sigma$         & 1.0              & 1.0              \\
$\lambda$    & 1.0               & 1.0          \\
$\gamma$         & 30            & 15              \\
Learning rate & 0.1        & 0.1            \\ \bottomrule
\end{tabular}
\caption{\textbf{Optimal Setting.} Representation of different setting considering hyperparameters for TransEAML (the rest of the models have been trained with their best settings in their own original resources).}
\label{tab:3}
\end{table}


\subsection{Experimental Setup} 
The TransE model as well as our proposed loss functions can be trained with different settings on hyperparameters. 
For TransE, embedding dimension ($d$) and a number of generated negative samples ($n$) per each positive are selected as the two hyerparameters. 
Adaptive Margin Loss (AML) has $\gamma$, $\lambda_+$, $\lambda_-$ and $\sigma$ as hyperparameters.
TransE which is trained by margin ranking loss, Limited-Score Loss, soft margin loss and adaptive margin loss are denoted by TransE, TransE-RS, TransE-SM and TransEAML respectively. 
The implementation of TransEAML has been done in Pytorch using Adam and Adagrad as optimizers. 
The model stops training when the accuracy of hit@10 reaches a pick value and starts to grade down.  

Batch sizes of $512$ and $1024$ are tested for each dataset. 
In order to investigate the core effectiveness of the proposed loss function and have a fair comparison, embedding dimension is set to 100 (\autoref{tab:1}). 
Moreover, only one negative sample is generated per each positive sample. 
To reduce the number of parameters for searching, we set $\lambda_+, \lambda_-$ to $1$.
$\gamma$ and $\sigma$ are tuned in the sets ${0, 1, 2, 3, 4, 5, 10, 15, 20, 25, 30}$ and ${0.01, 0.1, 1, 10, 100}$ respectively. 
The optimal hyperparameters obtained for each dataset are reported in the \autoref{tab:3}. 
The experimental datasets of evaluation includes FB15k and WN18. 

\subsection{Results and Discussion}
The results represented in \autoref{tab:1} shows comparisons of TransEAML with TransE-RS, TransH-RS, TransE and TransH. 
Additionally, we compare our model to LMF, SME, SE, RESCAL and UNSTRUCTURES. 
To have a fair comparison to the models, we set $d = 100$ and only one negative sample is generated per each positive one. 
 \textit{unif} refers to the uniform negative sampling in which probability of corruption of head ($?,r,t$) or tail ($h,r,?$) are same. 
The \textit{bern} negative sampling \cite{wang2014knowledge} considers different probabilities for head ($?,r,t$) and tail ($h,r,?$) corruptions to reduce number of false negative samples. 
 Results reported in \autoref{tab:1} for other models are taken from their original publication of research works. 
 However, we re-implemented TransE with soft margin loss (TransE-SM). 

According to the results, TransE which is trained by MRL gets 89.2 and 47.1 on WN18 and FB15K respectively. 
TransE-RS which is trained by the limited-based score loss improves the results on both of the datasets. 
It gets 93.7 and 72.3 on WN18 and FB15K respectively. The results confirm that adding the term $[f_r(h,t) - \gamma_1]_+$ to the MRL significantly improves the performance of TransE model. 
TransEAML obtains 95.2 and 77.7 on WN18 and FB15K. 
Therefore, the proposed loss function improves the accuracy of TransE. 


\section{Conclusion} 
\label{sec:concl}

To improve the performance knowledge graph embedding models, we propose the Adaptive Margin Loss (AML) to tackle the problem of obtaining a margin automatically during the training process. 
In contrast to other approaches which are using manual settings for the upper and lower bound of positive and negative samples within a large search space, AML adapts the center of the margin.
Therefore, by adding a slack variable of the same value to the side of positive and negative samples, upper and lower bounds are determined automatically. 
TransEAML, TransE trained by Adaptive Margin Loss (AML), is evaluated in terms of mean rank and hit@10 of the other loss functions. 
The results approved a significant improve in accuracy with our proposed loss function.
TransEAML performs 95.2\% on filter of WN18 whereas TransE trained by Margin Ranking Loss is reported to be 89.2\% in Hits@10 and Limited-based Scoring Loss result is stated to have 93.7\% of accuracy. 
On FB15K, the difference is higher as TransEAML reaches 77.8\% while TransE on MRL is 47.1\% and 72.3\% is the reported accuracy for Limited-based Scoring.


%
\bibliographystyle{abbrv} 
\bibliography{main.bbl}


\begin{thebibliography}{29}


\ifx \showCODEN    \undefined \def \showCODEN     #1{\unskip}     \fi
\ifx \showDOI      \undefined \def \showDOI       #1{#1}\fi
\ifx \showISBNx    \undefined \def \showISBNx     #1{\unskip}     \fi
\ifx \showISBNxiii \undefined \def \showISBNxiii  #1{\unskip}     \fi
\ifx \showISSN     \undefined \def \showISSN      #1{\unskip}     \fi
\ifx \showLCCN     \undefined \def \showLCCN      #1{\unskip}     \fi
\ifx \shownote     \undefined \def \shownote      #1{#1}          \fi
\ifx \showarticletitle \undefined \def \showarticletitle #1{#1}   \fi
\ifx \showURL      \undefined \def \showURL       {\relax}        \fi
\providecommand\bibfield[2]{#2}
\providecommand\bibinfo[2]{#2}
\providecommand\natexlab[1]{#1}
\providecommand\showeprint[2][]{arXiv:#2}

\bibitem[\protect\citeauthoryear{Adams}{Adams}{2019}]%
        {adams2019surfing}
\bibfield{author}{\bibinfo{person}{Sam Adams}.}
  \bibinfo{year}{2019}\natexlab{}.
\newblock \showarticletitle{Surfing the Hype Cycle to Infinity and Beyond}.
\newblock \bibinfo{journal}{\emph{Research-Technology Management}}
  \bibinfo{volume}{62}, \bibinfo{number}{3} (\bibinfo{year}{2019}),
  \bibinfo{pages}{45--51}.
\newblock


\bibitem[\protect\citeauthoryear{Anonymous}{Anonymous}{2018}]%
        {anonymous2018relation}
\bibfield{author}{\bibinfo{person}{Anonymous}.}
  \bibinfo{year}{2018}\natexlab{}.
\newblock \bibinfo{title}{Relation Pattern Encoded Knowledge Graph Embedding by
  Translating in Complex Space}.  (\bibinfo{year}{2018}).
\newblock
\newblock
\shownote{anonymous preprint under review.}


\bibitem[\protect\citeauthoryear{Bini}{Bini}{2018}]%
        {bini2018artificial}
\bibfield{author}{\bibinfo{person}{Stefano~A Bini}.}
  \bibinfo{year}{2018}\natexlab{}.
\newblock \showarticletitle{Artificial intelligence, machine learning, deep
  learning, and cognitive computing: what do these terms mean and how will they
  impact health care?}
\newblock \bibinfo{journal}{\emph{The Journal of arthroplasty}}
  \bibinfo{volume}{33}, \bibinfo{number}{8} (\bibinfo{year}{2018}),
  \bibinfo{pages}{2358--2361}.
\newblock


\bibitem[\protect\citeauthoryear{Bollacker, Evans, Paritosh, Sturge, and
  Taylor}{Bollacker et~al\mbox{.}}{2008}]%
        {bollacker2008freebase}
\bibfield{author}{\bibinfo{person}{Kurt Bollacker}, \bibinfo{person}{Colin
  Evans}, \bibinfo{person}{Praveen Paritosh}, \bibinfo{person}{Tim Sturge},
  {and} \bibinfo{person}{Jamie Taylor}.} \bibinfo{year}{2008}\natexlab{}.
\newblock \showarticletitle{Freebase: a collaboratively created graph database
  for structuring human knowledge}. In \bibinfo{booktitle}{\emph{Proceedings of
  the 2008 ACM SIGMOD international conference on Management of data}}. AcM,
  \bibinfo{pages}{1247--1250}.
\newblock


\bibitem[\protect\citeauthoryear{Bordes, Chopra, and Weston}{Bordes
  et~al\mbox{.}}{2014}]%
        {bordes2014question}
\bibfield{author}{\bibinfo{person}{Antoine Bordes}, \bibinfo{person}{Sumit
  Chopra}, {and} \bibinfo{person}{Jason Weston}.}
  \bibinfo{year}{2014}\natexlab{}.
\newblock \showarticletitle{Question answering with subgraph embeddings}.
\newblock \bibinfo{journal}{\emph{arXiv preprint arXiv:1406.3676}}
  (\bibinfo{year}{2014}).
\newblock


\bibitem[\protect\citeauthoryear{Bordes, Glorot, Weston, and Bengio}{Bordes
  et~al\mbox{.}}{2012}]%
        {bordes2012joint}
\bibfield{author}{\bibinfo{person}{Antoine Bordes}, \bibinfo{person}{Xavier
  Glorot}, \bibinfo{person}{Jason Weston}, {and} \bibinfo{person}{Yoshua
  Bengio}.} \bibinfo{year}{2012}\natexlab{}.
\newblock \showarticletitle{Joint learning of words and meaning representations
  for open-text semantic parsing}. In \bibinfo{booktitle}{\emph{Artificial
  Intelligence and Statistics}}. \bibinfo{pages}{127--135}.
\newblock


\bibitem[\protect\citeauthoryear{Bordes, Usunier, Garcia-Duran, Weston, and
  Yakhnenko}{Bordes et~al\mbox{.}}{2013}]%
        {bordes2013translating}
\bibfield{author}{\bibinfo{person}{Antoine Bordes}, \bibinfo{person}{Nicolas
  Usunier}, \bibinfo{person}{Alberto Garcia-Duran}, \bibinfo{person}{Jason
  Weston}, {and} \bibinfo{person}{Oksana Yakhnenko}.}
  \bibinfo{year}{2013}\natexlab{}.
\newblock \showarticletitle{Translating embeddings for modeling
  multi-relational data}. In \bibinfo{booktitle}{\emph{Advances in neural
  information processing systems}}. \bibinfo{pages}{2787--2795}.
\newblock


\bibitem[\protect\citeauthoryear{Bordes, Weston, Collobert, and Bengio}{Bordes
  et~al\mbox{.}}{2011}]%
        {bordes2011learning}
\bibfield{author}{\bibinfo{person}{Antoine Bordes}, \bibinfo{person}{Jason
  Weston}, \bibinfo{person}{Ronan Collobert}, {and} \bibinfo{person}{Yoshua
  Bengio}.} \bibinfo{year}{2011}\natexlab{}.
\newblock \showarticletitle{Learning structured embeddings of knowledge bases}.
  In \bibinfo{booktitle}{\emph{Twenty-Fifth AAAI Conference on Artificial
  Intelligence}}.
\newblock


\bibitem[\protect\citeauthoryear{Boyd and Vandenberghe}{Boyd and
  Vandenberghe}{2004}]%
        {boyd2004convex}
\bibfield{author}{\bibinfo{person}{Stephen Boyd} {and} \bibinfo{person}{Lieven
  Vandenberghe}.} \bibinfo{year}{2004}\natexlab{}.
\newblock \bibinfo{booktitle}{\emph{Convex optimization}}.
\newblock \bibinfo{publisher}{Cambridge university press}.
\newblock


\bibitem[\protect\citeauthoryear{Carlson, Betteridge, Kisiel, Settles,
  Hruschka, and Mitchell}{Carlson et~al\mbox{.}}{2010}]%
        {carlson2010toward}
\bibfield{author}{\bibinfo{person}{Andrew Carlson}, \bibinfo{person}{Justin
  Betteridge}, \bibinfo{person}{Bryan Kisiel}, \bibinfo{person}{Burr Settles},
  \bibinfo{person}{Estevam~R Hruschka}, {and} \bibinfo{person}{Tom~M
  Mitchell}.} \bibinfo{year}{2010}\natexlab{}.
\newblock \showarticletitle{Toward an architecture for never-ending language
  learning}. In \bibinfo{booktitle}{\emph{Twenty-Fourth AAAI Conference on
  Artificial Intelligence}}.
\newblock


\bibitem[\protect\citeauthoryear{Chen, Sun, Tu, Lin, and Liu}{Chen
  et~al\mbox{.}}{2016}]%
        {chen2016neural}
\bibfield{author}{\bibinfo{person}{Huimin Chen}, \bibinfo{person}{Maosong Sun},
  \bibinfo{person}{Cunchao Tu}, \bibinfo{person}{Yankai Lin}, {and}
  \bibinfo{person}{Zhiyuan Liu}.} \bibinfo{year}{2016}\natexlab{}.
\newblock \showarticletitle{Neural sentiment classification with user and
  product attention}. In \bibinfo{booktitle}{\emph{Proceedings of the 2016
  conference on empirical methods in natural language processing}}.
  \bibinfo{pages}{1650--1659}.
\newblock


\bibitem[\protect\citeauthoryear{He, Liu, Zhang, Xu, and Zhao}{He
  et~al\mbox{.}}{2014}]%
        {he2014question}
\bibfield{author}{\bibinfo{person}{Shizhu He}, \bibinfo{person}{Kang Liu},
  \bibinfo{person}{Yuanzhe Zhang}, \bibinfo{person}{Liheng Xu}, {and}
  \bibinfo{person}{Jun Zhao}.} \bibinfo{year}{2014}\natexlab{}.
\newblock \showarticletitle{Question answering over linked data using
  first-order logic}. In \bibinfo{booktitle}{\emph{Proceedings of the 2014
  Conference on Empirical Methods in Natural Language Processing (EMNLP)}}.
  \bibinfo{pages}{1092--1103}.
\newblock


\bibitem[\protect\citeauthoryear{H{\"o}ffner, Walter, Marx, Usbeck, Lehmann,
  and Ngomo}{H{\"o}ffner et~al\mbox{.}}{2016}]%
        {swj_survey_qa}
\bibfield{author}{\bibinfo{person}{Konrad H{\"o}ffner},
  \bibinfo{person}{Sebastian Walter}, \bibinfo{person}{Edgard Marx},
  \bibinfo{person}{Ricardo Usbeck}, \bibinfo{person}{Jens Lehmann}, {and}
  \bibinfo{person}{Axel-Cyrille~Ngonga Ngomo}.}
  \bibinfo{year}{2016}\natexlab{}.
\newblock \showarticletitle{Survey on Challenges of Question Answering in the
  Semantic Web}.
\newblock \bibinfo{journal}{\emph{Semantic Web Journal}}
  (\bibinfo{year}{2016}).
\newblock


\bibitem[\protect\citeauthoryear{Jenatton, Roux, Bordes, and
  Obozinski}{Jenatton et~al\mbox{.}}{2012}]%
        {jenatton2012latent}
\bibfield{author}{\bibinfo{person}{Rodolphe Jenatton},
  \bibinfo{person}{Nicolas~L Roux}, \bibinfo{person}{Antoine Bordes}, {and}
  \bibinfo{person}{Guillaume~R Obozinski}.} \bibinfo{year}{2012}\natexlab{}.
\newblock \showarticletitle{A latent factor model for highly multi-relational
  data}. In \bibinfo{booktitle}{\emph{Advances in Neural Information Processing
  Systems}}. \bibinfo{pages}{3167--3175}.
\newblock


\bibitem[\protect\citeauthoryear{Ji, He, Xu, Liu, and Zhao}{Ji
  et~al\mbox{.}}{2015}]%
        {ji2015knowledge}
\bibfield{author}{\bibinfo{person}{Guoliang Ji}, \bibinfo{person}{Shizhu He},
  \bibinfo{person}{Liheng Xu}, \bibinfo{person}{Kang Liu}, {and}
  \bibinfo{person}{Jun Zhao}.} \bibinfo{year}{2015}\natexlab{}.
\newblock \showarticletitle{Knowledge graph embedding via dynamic mapping
  matrix}. In \bibinfo{booktitle}{\emph{Proceedings of the 53rd Annual Meeting
  of the Association for Computational Linguistics and the 7th International
  Joint Conference on Natural Language Processing (Volume 1: Long Papers)}},
  Vol.~\bibinfo{volume}{1}. \bibinfo{pages}{687--696}.
\newblock


\bibitem[\protect\citeauthoryear{Lehmann, Isele, Jakob, Jentzsch, Kontokostas,
  Mendes, Hellmann, Morsey, van Kleef, Auer, and Bizer}{Lehmann
  et~al\mbox{.}}{2015}]%
        {swj_dbpedia}
\bibfield{author}{\bibinfo{person}{Jens Lehmann}, \bibinfo{person}{Robert
  Isele}, \bibinfo{person}{Max Jakob}, \bibinfo{person}{Anja Jentzsch},
  \bibinfo{person}{Dimitris Kontokostas}, \bibinfo{person}{Pablo Mendes},
  \bibinfo{person}{Sebastian Hellmann}, \bibinfo{person}{Mohamed Morsey},
  \bibinfo{person}{Patrick van Kleef}, \bibinfo{person}{S{\"o}ren Auer}, {and}
  \bibinfo{person}{Chris Bizer}.} \bibinfo{year}{2015}\natexlab{}.
\newblock \showarticletitle{{DB}pedia - A Large-scale, Multilingual Knowledge
  Base Extracted from Wikipedia}.
\newblock \bibinfo{journal}{\emph{Semantic Web Journal}} \bibinfo{volume}{6},
  \bibinfo{number}{2} (\bibinfo{year}{2015}), \bibinfo{pages}{167--195}.
\newblock
\newblock
\shownote{Outstanding Paper Award (Best 2014 SWJ Paper).}


\bibitem[\protect\citeauthoryear{Lin, Liu, Sun, Liu, and Zhu}{Lin
  et~al\mbox{.}}{2015}]%
        {lin2015learning}
\bibfield{author}{\bibinfo{person}{Yankai Lin}, \bibinfo{person}{Zhiyuan Liu},
  \bibinfo{person}{Maosong Sun}, \bibinfo{person}{Yang Liu}, {and}
  \bibinfo{person}{Xuan Zhu}.} \bibinfo{year}{2015}\natexlab{}.
\newblock \showarticletitle{Learning entity and relation embeddings for
  knowledge graph completion}. In \bibinfo{booktitle}{\emph{Twenty-ninth AAAI
  conference on artificial intelligence}}.
\newblock


\bibitem[\protect\citeauthoryear{Lin, Shen, Liu, Luan, and Sun}{Lin
  et~al\mbox{.}}{2016}]%
        {lin2016neural}
\bibfield{author}{\bibinfo{person}{Yankai Lin}, \bibinfo{person}{Shiqi Shen},
  \bibinfo{person}{Zhiyuan Liu}, \bibinfo{person}{Huanbo Luan}, {and}
  \bibinfo{person}{Maosong Sun}.} \bibinfo{year}{2016}\natexlab{}.
\newblock \showarticletitle{Neural relation extraction with selective attention
  over instances}. In \bibinfo{booktitle}{\emph{Proceedings of the 54th Annual
  Meeting of the Association for Computational Linguistics (Volume 1: Long
  Papers)}}, Vol.~\bibinfo{volume}{1}. \bibinfo{pages}{2124--2133}.
\newblock


\bibitem[\protect\citeauthoryear{Liu, Pokharel, and Pr{\'\i}ncipe}{Liu
  et~al\mbox{.}}{2007}]%
        {liu2007correntropy}
\bibfield{author}{\bibinfo{person}{Weifeng Liu}, \bibinfo{person}{Puskal~P
  Pokharel}, {and} \bibinfo{person}{Jos{\'e}~C Pr{\'\i}ncipe}.}
  \bibinfo{year}{2007}\natexlab{}.
\newblock \showarticletitle{Correntropy: Properties and applications in
  non-Gaussian signal processing}.
\newblock \bibinfo{journal}{\emph{IEEE Transactions on Signal Processing}}
  \bibinfo{volume}{55}, \bibinfo{number}{11} (\bibinfo{year}{2007}),
  \bibinfo{pages}{5286--5298}.
\newblock


\bibitem[\protect\citeauthoryear{Miller}{Miller}{1995}]%
        {miller1995wordnet}
\bibfield{author}{\bibinfo{person}{George~A Miller}.}
  \bibinfo{year}{1995}\natexlab{}.
\newblock \showarticletitle{WordNet: a lexical database for English}.
\newblock \bibinfo{journal}{\emph{Commun. ACM}} \bibinfo{volume}{38},
  \bibinfo{number}{11} (\bibinfo{year}{1995}), \bibinfo{pages}{39--41}.
\newblock


\bibitem[\protect\citeauthoryear{Nayyeri, Vahdati, Lehmann, and Yazdi}{Nayyeri
  et~al\mbox{.}}{2019}]%
        {nayyeri2019soft}
\bibfield{author}{\bibinfo{person}{Mojtaba Nayyeri}, \bibinfo{person}{Sahar
  Vahdati}, \bibinfo{person}{Jens Lehmann}, {and}
  \bibinfo{person}{Hamed~Shariat Yazdi}.} \bibinfo{year}{2019}\natexlab{}.
\newblock \showarticletitle{Soft Marginal TransE for Scholarly Knowledge Graph
  Completion}.
\newblock \bibinfo{journal}{\emph{arXiv preprint arXiv:1904.12211}}
  (\bibinfo{year}{2019}).
\newblock


\bibitem[\protect\citeauthoryear{Nickel, Tresp, and Kriegel}{Nickel
  et~al\mbox{.}}{2012}]%
        {nickel2012factorizing}
\bibfield{author}{\bibinfo{person}{Maximilian Nickel}, \bibinfo{person}{Volker
  Tresp}, {and} \bibinfo{person}{Hans-Peter Kriegel}.}
  \bibinfo{year}{2012}\natexlab{}.
\newblock \showarticletitle{Factorizing yago: scalable machine learning for
  linked data}. In \bibinfo{booktitle}{\emph{Proceedings of the 21st
  international conference on World Wide Web}}. ACM, \bibinfo{pages}{271--280}.
\newblock


\bibitem[\protect\citeauthoryear{Panetta}{Panetta}{5}]%
        {panetta5trends}
\bibfield{author}{\bibinfo{person}{Kasey Panetta}.}
  \bibinfo{year}{5}\natexlab{}.
\newblock \showarticletitle{trends emerge in the gartner hype cycle for
  emerging technologies, 2018}.
\newblock \bibinfo{journal}{\emph{Retrieved November}}  \bibinfo{volume}{4}
  (\bibinfo{year}{5}), \bibinfo{pages}{2018}.
\newblock


\bibitem[\protect\citeauthoryear{Shen, Wang, Luo, and Wang}{Shen
  et~al\mbox{.}}{2013}]%
        {shen2013linking}
\bibfield{author}{\bibinfo{person}{Wei Shen}, \bibinfo{person}{Jianyong Wang},
  \bibinfo{person}{Ping Luo}, {and} \bibinfo{person}{Min Wang}.}
  \bibinfo{year}{2013}\natexlab{}.
\newblock \showarticletitle{Linking named entities in tweets with knowledge
  base via user interest modeling}. In \bibinfo{booktitle}{\emph{Proceedings of
  the 19th ACM SIGKDD international conference on Knowledge discovery and data
  mining}}. ACM, \bibinfo{pages}{68--76}.
\newblock


\bibitem[\protect\citeauthoryear{Szumlanski and Gomez}{Szumlanski and
  Gomez}{2010}]%
        {szumlanski2010automatically}
\bibfield{author}{\bibinfo{person}{Sean Szumlanski} {and}
  \bibinfo{person}{Fernando Gomez}.} \bibinfo{year}{2010}\natexlab{}.
\newblock \showarticletitle{Automatically acquiring a semantic network of
  related concepts}. In \bibinfo{booktitle}{\emph{Proceedings of the 19th ACM
  international conference on Information and knowledge management}}. ACM,
  \bibinfo{pages}{19--28}.
\newblock


\bibitem[\protect\citeauthoryear{Trouillon, Welbl, Riedel, Gaussier, and
  Bouchard}{Trouillon et~al\mbox{.}}{2016}]%
        {trouillon2016complex}
\bibfield{author}{\bibinfo{person}{Th{\'e}o Trouillon},
  \bibinfo{person}{Johannes Welbl}, \bibinfo{person}{Sebastian Riedel},
  \bibinfo{person}{{\'E}ric Gaussier}, {and} \bibinfo{person}{Guillaume
  Bouchard}.} \bibinfo{year}{2016}\natexlab{}.
\newblock \showarticletitle{Complex embeddings for simple link prediction}. In
  \bibinfo{booktitle}{\emph{International Conference on Machine Learning}}.
  \bibinfo{pages}{2071--2080}.
\newblock


\bibitem[\protect\citeauthoryear{Wang, Zhang, Feng, and Chen}{Wang
  et~al\mbox{.}}{2014}]%
        {wang2014knowledge}
\bibfield{author}{\bibinfo{person}{Zhen Wang}, \bibinfo{person}{Jianwen Zhang},
  \bibinfo{person}{Jianlin Feng}, {and} \bibinfo{person}{Zheng Chen}.}
  \bibinfo{year}{2014}\natexlab{}.
\newblock \showarticletitle{Knowledge graph embedding by translating on
  hyperplanes}. In \bibinfo{booktitle}{\emph{Twenty-Eighth AAAI conference on
  artificial intelligence}}.
\newblock


\bibitem[\protect\citeauthoryear{Zhang, Yuan, Lian, Xie, and Ma}{Zhang
  et~al\mbox{.}}{2016}]%
        {zhang2016collaborative}
\bibfield{author}{\bibinfo{person}{Fuzheng Zhang},
  \bibinfo{person}{Nicholas~Jing Yuan}, \bibinfo{person}{Defu Lian},
  \bibinfo{person}{Xing Xie}, {and} \bibinfo{person}{Wei-Ying Ma}.}
  \bibinfo{year}{2016}\natexlab{}.
\newblock \showarticletitle{Collaborative knowledge base embedding for
  recommender systems}. In \bibinfo{booktitle}{\emph{Proceedings of the 22nd
  ACM SIGKDD international conference on knowledge discovery and data mining}}.
  ACM, \bibinfo{pages}{353--362}.
\newblock


\bibitem[\protect\citeauthoryear{Zhou, Zhu, Liu, and Guo}{Zhou
  et~al\mbox{.}}{2017}]%
        {zhou2017learning}
\bibfield{author}{\bibinfo{person}{Xiaofei Zhou}, \bibinfo{person}{Qiannan
  Zhu}, \bibinfo{person}{Ping Liu}, {and} \bibinfo{person}{Li Guo}.}
  \bibinfo{year}{2017}\natexlab{}.
\newblock \showarticletitle{Learning knowledge embeddings by combining
  limit-based scoring loss}. In \bibinfo{booktitle}{\emph{Proceedings of the
  2017 ACM on Conference on Information and Knowledge Management}}. ACM,
  \bibinfo{pages}{1009--1018}.
\newblock


\end{thebibliography}

\end{document}